\documentclass[letterpaper, 10 pt, conference]{ieeeconf}  %

\IEEEoverridecommandlockouts                              %

\usepackage{amsmath} %
\usepackage{graphicx}
\usepackage{booktabs}
\usepackage{multirow}
\usepackage[table,dvipsnames]{xcolor}
\usepackage{tikz}
\usetikzlibrary{arrows.meta,calc,backgrounds}
\definecolor{dashorange}{HTML}{D55E00}
\definecolor{dashblue}{HTML}{0072B2}
\definecolor{dashink}{HTML}{263441}

\title{\LARGE \bf
DASH-SLAM: Calibration-Free Metric Monocular SLAM\\ via Depth Scale Propagation
}

\author{Otavio Ribas$^{1,2}$, Jacopo Panerati$^{3}$, and Pierre-Yves Lajoie$^{1,2}$%
\thanks{$^{1}$Department of Computer and Software Engineering, Polytechnique Montreal, Montreal, Canada.}%
\thanks{$^{2}$Mila - Quebec AI Institute, Montreal, Canada.}%
\thanks{$^{3}$National Research Council Canada, Montreal, Canada.}%
}

\makeatletter
\let\NAT@parse\undefined
\makeatother
\usepackage[numbers,sort&compress]{natbib}
\usepackage{url}

\begin{document}

\maketitle

\begin{abstract}

Geometric foundation models estimate dense geometry and camera parameters from monocular images, but reliable SLAM also requires consistent scale and accurate data association across successive reconstructions. We present DASH-SLAM, a monocular SLAM system that integrates metric depth predictions from pretrained Depth Anything~3 into a shared-keyframe pose graph without task-specific fine-tuning. DASH-SLAM derives scale corrections from depth ratios at shared anchor images and accumulates them across reconstructions, with a damping factor controlling their influence. These corrections rescale odometry translations, loop constraints, and map geometry. Experiments on TUM, Replica, and 7-Scenes show that DASH-SLAM outperforms state-of-the-art methods. We evaluate performance using both rigid and similarity alignment to distinguish metric-scale fidelity from accuracy up to scale. We further evaluate DASH-SLAM on four challenging sequences in perceptually-degraded environments from the Unified Autonomy Stack field dataset, covering tunnels, fog, and a frozen lake. On these sequences, DASH-SLAM outperforms the foundation-model SLAM state-of-the-art and maintains tracking where classical SLAM methods fail.

\end{abstract}

\section{INTRODUCTION}

Simultaneous Localization and Mapping (SLAM) is a foundational task in robotics to simultaneously produce a 3D representation of a scene and estimate the poses of the agent in that map. While LiDARs and other modern types of sensors are getting more common, a single RGB camera is still the cheapest, lightest, and most available sensor, which makes Monocular SLAM particularly attractive. However, this setting carries a set of constraints: many of the main existing systems, both classical and learned~\cite{orbslam3, droidslam, dpvslam}, rely on calibration, and a single camera is physically unable to observe absolute scale. Although these problems can sometimes be mitigated in a controlled environment, they pose major problems in real-world applications, where camera intrinsics are often unknown and the ambiguity on scale renders it impossible for a robot to measure the distance it has travelled.

Geometric Foundation Models (GFMs) have greatly contributed to overcome the first limitation, i.e. calibration, by being trained to estimate a point map, camera poses, and intrinsics directly from uncalibrated images~\cite{dust3r, mast3r, vggt, da3}. Consequently, SLAM systems that leverage these models as their backbone bypass calibration altogether, and even manage to surpass some calibrated pipelines on several benchmarks~\cite{mast3rslam, vggtslam2}, thus making it possible to run SLAM systems on arbitrary cameras.

The second limitation, i.e. scale, remains largely unresolved and is fundamentally more complicated to mitigate. In fact, this limitation is often concealed in the standard evaluation protocol: Monocular SLAM accuracy is generally measured by its absolute trajectory error after a $Sim(3)$ alignment to ground truth, which estimates and removes a global scale factor before any error is measured. In practice, this means that a map that is too large or too small is indistinguishable from a correctly scaled one. As almost none of the state-of-the-art monocular SLAM systems recovers metric scale, some methods that perform well under $\mathrm{Sim}(3)$ alignment can exhibit errors more than an order of magnitude larger when evaluated under rigid $\mathrm{SE}(3)$ alignment. Yet, accurate scale remains a requirement for robots to actually navigate real-world settings.

The latest GFMs such as Depth Anything~3 (DA3)~\cite{da3} have the capability of predicting depth in metric units, anchoring their reconstruction on scale priors learned from their real-world training data. However, harvesting this property into a proper metric SLAM system is not automatic. While some foundation models can operate over an arbitrarily large sequence of frames, they process them all at once. This means that simply feeding the frames to the model would not allow a robot to estimate its state online. Moreover, GFMs can reach memory constraints relatively fast, which is a real limitation in autonomous systems. This means a system based on a GFM is required to chain the estimations for each batch, but since each carries its own small scale error, these errors compound and the scale drifts. The question is how to use a metric GFM so that its scale estimates inform the whole sequence, without calibration and without returning to a hand-crafted feature tracking pipeline. While concurrent work injects per-keyframe DA3 depth priors into the back-end of a feature-based SLAM pipeline~\cite{prismslam}, we opted to use the foundation model's joint multi-view reconstruction, which provides poses, intrinsics, and metric depth, as the tracking front-end itself.

\begin{figure*}[t!]
\centering
\begin{tikzpicture}[
  font=\sffamily\fontsize{8}{10}\selectfont,
  >={Latex[length=1.6mm]},
  flow/.style={->,line width=.8pt,draw=dashink},
  box/.style={draw=dashink!35,fill=white,rounded corners=2pt,line width=.6pt},
  title/.style={font=\sffamily\bfseries\fontsize{8}{10}\selectfont,text=dashink},
  small/.style={font=\sffamily\fontsize{7}{9}\selectfont,text=dashink!85},
]
\path[box,fill=dashblue!4] (0,0) rectangle (3.1,1.65);
\node[title] at (1.55,1.35) {Select RGB keyframes};
\node[small] at (1.55,1.04) {Shared anchors in orange};
\foreach \x/\lab in {.35/1,.83/2,1.31/3,1.79/4} {
  \draw[draw=dashblue,fill=dashblue!12] (\x,.56) rectangle +(0.34,.28);
}
\foreach \x in {1.31,1.79,2.27,2.75} {
  \draw[draw=dashblue,fill=dashblue!12] (\x,.13) rectangle +(0.34,.28);
}
\foreach \x in {1.31,1.79} {
  \fill[fill=dashorange!75] (\x,.56) rectangle +(0.34,.28);
  \fill[fill=dashorange!75] (\x,.13) rectangle +(0.34,.28);
  \draw[dashorange,dashed,line width=.6pt] (\x+.17,.54)--(\x+.17,.43);
}
\node[small,anchor=west] at (.22,.26) {Next};

\path[box,fill=dashblue!4] (3.55,0) rectangle (6.25,1.65);
\node[title] at (4.9,1.35) {Depth Anything 3};
\node[align=center,text=dashink,font=\sffamily\fontsize{7}{9}\selectfont] at (4.9,.68) {Joint inference of\\poses, intrinsics\\and metric depth};

\path[box,draw=dashorange,fill=dashorange!5,line width=.9pt] (6.7,0) rectangle (11.25,1.65);
\node[title,text=dashorange] at (8.975,1.35) {Anchor-depth scale chaining};
\node[text=dashink] at (8.975,.86) 
{$r_m={median}_{\mathbf p}\!\bigl[D_a^{m-1}(\mathbf p)/D_a^m(\mathbf p)\bigr]$};
\node[text=dashorange,font=\fontsize{10}{12}\selectfont] at (8.975,.34) {$S_m=S_{m-1}\,r_m^{\,1-g}$};

\path[box,fill=dashblue!4] (11.7,0) rectangle (14.55,1.65);
\node[title] at (13.125,1.35) {Graph optimization};
\foreach \x/\y in {12.0/.66,12.6/.48,13.25/.70,14.15/.51} {
  \fill[fill=dashblue] (\x,\y) circle (.075);
}
\draw[dashblue,line width=.8pt] (12,.66)--(12.6,.48)--(13.25,.70)--(14.15,.51);
\draw[dashorange,line width=.8pt] (12,.66) to[bend left=28] (14.15,.51);
\node[small] at (13.125,.17) {Shared anchor nodes};

\path[box] (15,0) rectangle (17.2,1.65);
\node[title] at (16.1,1.35) {Metric outputs};
\node[align=center,text=dashink,font=\sffamily\fontsize{7}{9}\selectfont] at (16.1,.69) {Trajectory\\[4pt]Dense point map};
\foreach \a/\b in {3.1/3.55,6.25/6.7,11.25/11.7,14.55/15} {
  \draw[flow] (\a,.825)--(\b,.825);
}
\node[box,minimum width=2.35cm,minimum height=.52cm,align=center] (retrieve) at (3.0,-.77) {Retrieve revisits};
\node[box,minimum width=2.8cm,minimum height=.52cm,align=center] (reinfer) at (7.05,-.77) {DA3 re-inference + checks};
\node[box,draw=dashorange,fill=dashorange!5,minimum width=2.75cm,minimum height=.52cm,align=center] (loop) at (11.36,-.77) {Scale loop translation};
\draw[flow] (1.55,0)--(1.55,-.77)--(retrieve.west);
\draw[flow] (retrieve)--(reinfer);
\draw[flow] (reinfer)--(loop);
\draw[flow] (loop.east)--(13.125,-.77)--(13.125,0);
\end{tikzpicture}
\caption{DASH-SLAM overview. The front end selects keyframes from the incoming image stream, and the inference stage constructs submaps using Depth Anything~3. The back end propagates metric scale across consecutive submaps through their shared anchor frames. The loop-closure thread detects revisited places and passes loop constraints to the back end, which optimises the global pose graph.}
\label{fig:pipeline}
\end{figure*}
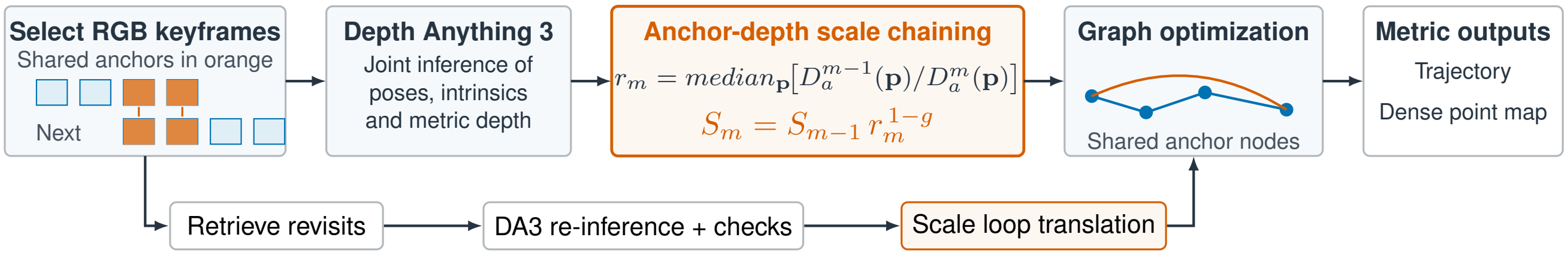

We present \textbf{DASH-SLAM} (\textbf{D}epth-\textbf{A}ssisted \textbf{S}cale-\textbf{H}armonized Metric Monocular SLAM), an uncalibrated RGB SLAM system that uses DA3 as its backbone for tracking, mapping, and loop closure. A batch of keyframes is processed by DA3 into poses and metric depth, which are then used to produce a submap. Consecutive submaps share two anchor frames that are observed in both inferences, allowing two independent depth maps to directly measure the relationship between the scale of two consecutive maps. DASH-SLAM chains these per-boundary depth ratios into an accumulated scale that expresses every submap in a single metric unit, and applies this scale to the translation of every relative-pose measurement entering the pose graph, while keeping rotations untouched. Because the scale ratio noise compounds over many boundaries, each ratio $r$ enters the chain as $r^{1-g}$, where $g$ represents a damping factor, ranging from fully chaining the scales ($g=0$) to solely relying on the metric scale of the foundation model ($g=1$). Loop closures are verified by re-running the foundation model on the matched frames and their temporal neighbors, closing cycles that constrain the chained scale. In summary, our main contributions are: 

\begin{itemize}
  \item \textbf{DASH-SLAM}, a calibration-free monocular SLAM system that uses a metric geometric foundation model as its tracking front-end, jointly reconstructing keyframe submaps, connecting them through shared anchor frames in a pose graph. To improve robustness, our system also integrates motion-adjusted keyframe selection and  loop closure verification by foundation-model re-inference. Our system achieves state-of-the-art $SE(3)$ and $Sim(3)$ accuracy among uncalibrated monocular systems on TUM RGB-D, Replica, and 7-Scenes.
  \item An \textbf{anchor-frame scale chaining} mechanism that propagates the foundation model's per-batch metric scale to odometry translations, loop constraints,
  and map geometry, with a damping factor controlling their influence. 
  \item \textbf{Evaluation beyond simple indoor benchmarks.} We evaluate DASH-SLAM on challenging field sequences from the Unified Autonomy Stack dataset~\citep{uas}. Complementing standard indoor benchmarks, this evaluation measures trajectory accuracy and completion rates to assess how well these methods generalize to repetitive and visually degraded environments, which remain particularly challenging for SLAM.
\end{itemize}

\section{RELATED WORKS}

\noindent\textbf{Visual SLAM and learned geometry.}
Feature-based systems such as ORB-SLAM3~\citep{orbslam3} estimate motion and structure through correspondence search and geometric optimization. Early learned systems retain this coupling: DROID-SLAM~\citep{droidslam} iteratively refines dense correspondences, camera poses, and depth through differentiable bundle adjustment, while DPV-SLAM~\citep{dpvslam} uses a sparse patch graph and efficient proximity loop closure. Their standard monocular formulations use camera calibration and do not determine absolute scale from geometry alone. DASH-SLAM instead obtains poses and depth jointly from a pretrained reconstruction model, with metric scale supplied by its learned prior.

\noindent\textbf{Foundation models for SLAM.}
In recent years, new machine learning architectures capable of learning from internet-scale datasets have had a major impact on computer vision. DUSt3R~\citep{dust3r} predicts pointmaps from uncalibrated image pairs, MASt3R~\citep{mast3r} adds learned matching features, and VGGT~\citep{vggt} jointly predicts geometry and cameras across multiple views. These priors support different sequential formulations useful for robotics. MASt3R-SLAM~\citep{mast3rslam} combines two-view pointmap matching with tracking, fusion, and global optimization; ViSTA-SLAM~\citep{vistaslam} learns symmetric two-view geometry and relative poses for a $\mathrm{Sim}(3)$ graph. 
A different approach, FLOCC-SLAM~\citep{floccslam}, uses GFM to estimate inter-robot loop closures and optimizes their scales against existing metric odometry. DASH-SLAM instead obtains both odometry and loop measurements from DA3, whose metric scale comes from the model's learned prior.

More closely related to our approach are VGGT-SLAM~\citep{vggtslam}, which aligns overlapping submaps on $\mathrm{SL}(4)$, and its successor, VGGT-SLAM~2.0~\citep{vggtslam2}, which uses keyframe nodes, overlap constraints that reconcile calibration and scale, and re-inferred loop submaps. 
While these approaches target up-to-scale reconstruction, DASH-SLAM leverages DA3's metric priors to merge repeated anchor images into shared nodes, and applies depth-derived scale corrections with damping, to ensure metric accuracy and better consistency along the robot's trajectory. 

A related approach, using the same inference backbone as ours,
DA3-Streaming~\citep{da3streaming}, reconstructs overlapping image batches and aligns them through similarity transformations derived from shared geometry and loop closures, repeatedly processing the overlapping frames. In contrast, DASH-SLAM speeds up the pipeline using smaller overlaps and optimizes individual keyframe poses through scale-adjusted relative-pose constraints, enabling corrections within and across submaps.

\noindent\textbf{Metric reconstruction priors.}
Metric depth estimation has a long history in computer vision, and recent approaches like Metric3D~\citep{metric3d} and UniDepth~\citep{unidepth} can learn metric geometry across camera settings. MapAnything~\citep{mapanything} and
Depth Anything~3~\citep{da3} extend the available reconstruction priors to multi-view settings.

Closely related to our approach, AMB3R~\citep{amb3r} combines a learned metric-scale head with a compact volumetric backend and introduces AMB3R-VO, a model-agnostic odometry pipeline that also supports DA3. In comparison, DASH-SLAM retains an explicit pose graph with retrieval-based loop constraints for correcting the trajectory. Another technique PRISM-SLAM~\citep{prismslam} incorporates DA3 depth priors into a feature-tracking system using uncertainty weighting, scale filtering, and metric optimization. DASH-SLAM uses instead joint multi-view inference itself to obtain relative poses, then converts both odometry and re-inferred loop measurements into a common scale. Thus, one of our contributions concerns the integration and propagation of metric predictions within SLAM. We assess this integration under
both $\mathrm{SE}(3)$ and $\mathrm{Sim}(3)$ alignments.

\noindent\textbf{Field robustness and perceptual aliasing.}
Perceptual aliasing can produce correlated false loop constraints that can lead to drastic failures, a longstanding challenge in SLAM~\citep{dcgm}. VGGT-SLAM~2.0~\citep{vggtslam2} studies retrieval errors in repetitive offices and demonstrates outdoor driving reconstruction. CoMo3R-SLAM~\citep{como3rslam} addresses outdoor cross-agent association, while On-the-Fly3R~\citep{onthefly3r} evaluates large UAV reconstructions and validates local estimates before integration. Our experiments complement these efforts with tunnel, fog, and frozen-lake sequences, examining the sensitivity, benefits and vulnerability of current metric-prior SLAM under challenging conditions, largely unsolved to this day.

\section{METHODOLOGY}
\label{sec:method}

DASH-SLAM takes a stream of monocular RGB frames with unknown intrinsics and produces a metric camera trajectory and a dense point map. It comprises of four parallel threads connected by bounded queues (see the overall architecture in Fig.~\ref{fig:pipeline}): a \emph{front-end} that selects keyframes, an \emph{inference} stage that builds submaps with Depth Anything~3 (DA3)~\cite{da3} by consuming selected keyframes, a \emph{back-end} that places each submap in a global metric frame and optimizes a pose graph, and a \emph{loop-closure} worker that detects, revisits, and measures them with the same GFM. We describe each stage in its dedicated subsection.

\subsection{Keyframe Selection}
\label{sec:method:keyframes}

Typically, visual SLAM systems select a frame as a keyframe when its disparity with respect to the previous keyframe exceeds a threshold~\cite{mast3r, vggtslam2}. In learning-based dense SLAM, keyframes are the input set for the feed-forward reconstruction. Therefore, a disparity-based approach causes the density of the input set to depend on scene depth and camera motion. 
Furthermore, since DA3 operates over a batch of keyframes, irregular density translates into batches that span too little motion to provide parallax, or too much to retain the view overlap its cross-view attention relies on. Therefore, instead of relying on frame-by-frame disparity, we select keyframes at the segment level~\cite{choi2026segment}: the stream is partitioned into non-overlapping segments of $N_S = 128$ frames, and keyframes are sampled regularly within that segment with a stride that is dependent on the apparent motion. For consecutive frames $i, i{+}1$ in a segment, we estimate the mean displacement magnitude $\bar{d}_{i,i+1}$ through optical flow. The accumulated displacement $D_S = \sum_i \bar{d}_{i,i+1}$ is then used to select the stride.

\begin{equation}
  \sigma = \left\{
  \begin{array}{ll}
    8,  & D_S > \tau_S,\\
    16, & \mbox{otherwise},
  \end{array}\right.
\end{equation}%

Consequently, fast motion is sampled twice as densely as slow motion, while keyframe spacing remains regular within a segment.

\subsection{Submap Reconstruction}
\label{sec:method:submaps}

Keyframes are grouped into batches, each of which DA3 reconstructs jointly in a single forward pass. For a batch $\mathcal{B}_m = \{I_1, \dots, I_{n}\}$, DA3 predicts, for every frame $i$, a metric depth map $\mathbf{D}_i$, a per-pixel confidence $\mathbf{C}_i$, an intrinsic matrix $\mathbf{K}_i$, and a world-to-camera pose $\mathbf{E}_i^{(m)}$ expressed in a frame local to the batch. 

\noindent\textbf{Batch size.} While Depth Anything 3 can process an arbitrarily long sequence, this can often exceed a system's memory constraints and/or increase latency. Thus, we set a maximum batch size of $n_\text{max}$ keyframes. Furthermore, since short test sequences could otherwise yield a single submap (leaving no boundary to chain and no graph to optimize), the size follows a warm-up ramp: the first two submaps containing $n_\text{warm}$ and all subsequent ones $n_\text{max}$.%

\noindent\textbf{Anchor frames.} Consecutive batches overlap by two \emph{anchor frames}, which are reconstructed by both batches. Doing so serves two purposes: providing a link between submaps that can be used for aligning them; and computing a direct measurement of their relative scale.

\noindent\textbf{Confidence filtering.} DA3 confidences are normalized
to $[0,1]$ within each frame, and pixels below the 65th percentile of all normalized confidences in that batch are then discarded. Each remaining pixel $(u,v)$ is then back-projected by using each frame's own predicted intrinsics $\mathbf{K}$, producing a point map $\mathbf{x}$.

\begin{equation}
    \mathbf{x} = \mathbf{D}_i(u,v)\,\mathbf{K}_i^{-1}[u,v,1]^\top.
\end{equation} %

\subsection{Anchor-Frame Scale Chaining}
\label{sec:method:scale}

Although DA3 predicts depth in metric units, each batch carries its own small scale error, and nothing constrains two batches to agree. As a consequence, composing submaps naively produces scale discontinuities at every boundary. To correct this, we compute the relative scale ratio $r_m$ directly at the first shared anchor frame $a$, as it appears in both batches with depth maps $\mathbf{D}_a^{(m-1)}$ and $\mathbf{D}_a^{(m)}$:
\begin{equation}
  r_m = \mathop{\mathrm{median}}\limits_{\mathbf{p} \in \Omega_a}
        \frac{\mathbf{D}_a^{(m-1)}(\mathbf{p})}{\mathbf{D}_a^{(m)}(\mathbf{p})},
\end{equation}
where $\Omega_a$ is the set of pixels with finite, positive depth in both maps. Using the median makes the estimate robust to local depth failures in either batch. The per-boundary ratios are then accumulated into a global scale that expresses submap $m$ in the metric unit of the first submap,
\begin{equation}
  S_0 = 1, \qquad S_m = S_{m-1}\, r_m^{\,1-g},
  \label{eq:chain}
\end{equation}
where $g \in [0,1]$ is a damping factor: $g = 0$ applies every measured ratio
in full, and $g = 1$ ignores the ratios and relies on each batch's own metric
scale. The standard mode of operation uses full chaining ($g=0$). Furthermore, loop closures (Sec.~\ref{sec:method:loops}) close cycles in the graph that constrain the
accumulated scale.

$S_m$ is used to adjust \emph{translations} of every relative-pose measurement
derived from submap $m$ before it enters the pose graph. \emph{Rotations}, which
are scale-invariant, are left untouched. For a relative pose
$\mathbf{Z} = [\mathbf{R} \,|\, \mathbf{t}]$ measured in submap $m$, we write
$\mathbf{Z}^{S_m} = [\mathbf{R} \,|\, S_m\mathbf{t}]$.

\subsection{Pose Graph}
\label{sec:method:graph}

The global trajectory is estimated with a pose graph containing one node
$\mathbf{X}_i$ per keyframe, where $\mathbf{X}_i$ is the camera-to-world
transform of keyframe $i$. Nodes are parameterized on the projective group
SL(4), following VGGT-SLAM~\cite{vggtslam}.

\noindent\textbf{Initialisation.} The first submap is inserted at its DA3 poses
and its first keyframe is anchored with a tight prior, which fixes the reference coordinate frame.
Each later submap is placed through its first anchor frame $a$, whose global pose
is already optimized: every other frame $j$ of the submap is initialized as, 
\begin{equation}
  \mathbf{X}_j = \mathbf{X}_a\, \bigl(\mathbf{E}_a^{(m)} (\mathbf{E}_j^{(m)})^{-1}\bigr)^{S_m}.
\end{equation}

\noindent\textbf{Odometry factors.} Each submap contributes relative-pose constraints to the pose graph. A constraint between keyframes $i$ and $j$ is the relative pose predicted by DA3, $\mathbf{Z}_{ij} = \mathbf{E}_i^{(m)} (\mathbf{E}_j^{(m)})^{-1}$, with its translation scaled by $S_m$. Constraints are added for three kinds of keyframe pairs:
(i)~consecutive keyframes $(i, i{+}1)$;
(ii)~$n_\text{stride}$ keyframes apart within the same batch, $(i, i{+}n_\text{stride})$; and
(iii)~the anchor pair, which is measured independently by both batches and
therefore contributes to two different constraints between the same two nodes.
Because DA3 estimates a batch jointly, the $n_\text{stride}$ factors are consistent with the
consecutive chain, so they do not add new measurements. However, they serve to stiffen each submap, so that disagreements introduced by boundary and loop-closure factors are absorbed at submap boundaries instead of bending the interior of a submap. The duplicated anchor factors, on the other hand, give the graph redundancy at the frontier between two consecutive submaps.

\noindent\textbf{Optimization.} All factors use an isotropic noise model with
$\sigma = 0.05$. After every new submap, the complete graph is optimized with
Levenberg--Marquardt in GTSAM~\cite{gtsam}, initialized using the previous
solution.

\subsection{Boundary Verification and Repair}
\label{sec:method:boundary}

Each boundary is verified for consistency by using the anchor pair. We compare the relative pose between the two anchor frames as measured by submap $m{-}1$ and by submap $m$, and flag the boundary as broken when the two disagree by more than $15^\circ$ in rotation, or when the ratio of their translation magnitudes lies outside $[0.5, 2]$. When this happens, DA3 repairs the boundary by producing a third measurement using frames from both submaps, used to arbitrate between the two conflicting factors. 

\subsection{Loop Closure}
\label{sec:method:loops}

\noindent\textbf{Retrieval.} Every keyframe is described by a DINO-SALAD global
descriptor~\cite{salad}. When a submap is completed, each of its keyframes is
compared against all previous ones that are at least two submaps
away, using the L2 distance between descriptors. Pairs closer than $l_\text{diff}$ are
candidates; only the closest candidate per earlier submap is retained, and at
most one candidate is verified per query submap.

\noindent\textbf{Verification by re-inference.} We verify the validity of a loop-closure by re-running DA3 on a small batch formed by the query keyframe, the matched keyframe, and neighbors of each one of them. A candidate is rejected if the mean predicted confidence over the two matched frames is below $c_{gate}$. Otherwise, all frames of the new batch share one local frame, and the loop measurement is the relative pose between the two matched frames, $\mathbf{Z}_{\mathrm{loop}} = \mathbf{E}_q (\mathbf{E}_d)^{-1}$.

\noindent\textbf{Metric scale of the loop.} As with every DA3 inference, the verification comes with its own scale. Thus, to adjust it to the global scale, first it needs to be brought to the same scale as the query submap by using the median depth ratio between the query keyframe's original depth map and its depth in the re-inference (as in
Sec.~\ref{sec:method:scale}). Then, it can be brought into the global scale by using the query submap's accumulated scale $S_m$ (Eq.~\ref{eq:chain}).

\noindent\textbf{Acceptance.} Each measured loop is compared with the relative
pose the current graph predicts between the two frames, and accepted immediately
if the rotation disagreement is below $90^\circ$. Loops that fail this check are
not discarded but \emph{held}: if odometry breaks, the graph itself is wrong, so a correct loop can appear inconsistent. Held loops are re-evaluated whenever a new submap is processed, and accepted if a second held loop implies the same correction. Furthermore, a held loop that remains uncontradicted for five submaps is accepted on its own. Accepted loops enter the graph as between-factors with a Huber robust kernel ($k_\text{loop}$), so
an erroneous closure that passes all checks is down-weighted rather than warping
the map. Boundary repairs follow the same path but bypass the geometric check,
since their frames are adjacent by construction.

\subsection{Map Reconstruction}
\label{sec:method:map}

Geometry is stored per keyframe, in camera coordinates, as the confidence-filtered
back-projected points of Sec.~\ref{sec:method:submaps}. The global map is obtained
by expressing each keyframe's points in the global metric unit and transforming
them with the keyframe's optimised pose,
$\mathbf{x}^{w} = \mathbf{X}_i \,(S_m\, \mathbf{x})$, with anchor frames taken
from the first submap that contains them. Because points are re-projected from the
current poses rather than fused into a fixed cloud, corrections from loop closures
propagate to the map without re-running the foundation model.

\begin{figure}[t]
    \centering
    \includegraphics[width=\columnwidth]{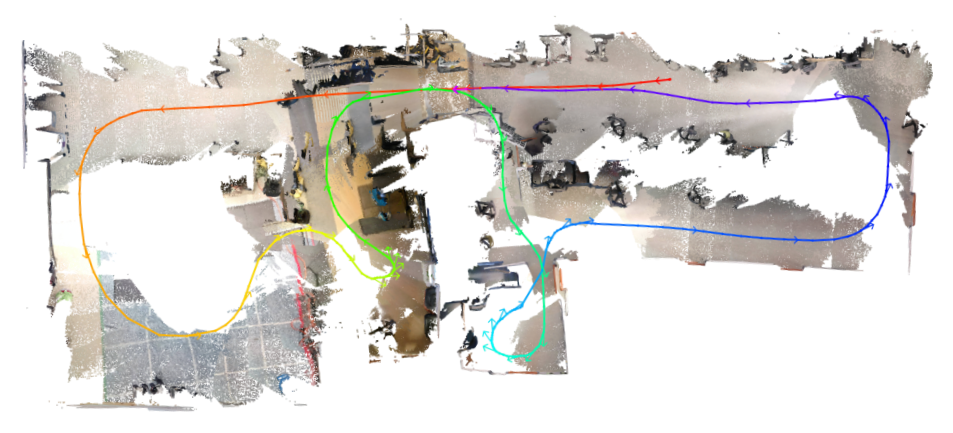}
    \caption{Visualization of a robot's trajectory and map obtained online with DASH-SLAM in a laboratory setting. Submaps are identified in different colors. In this real-world experiment, DASH-SLAM achieved 13.57 FPS and a peak RAM usage of 14.51 GB on an NVIDIA Jetson AGX Orin.}
    \label{fig:lab_map}
\end{figure}

\begin{table*}[t]
\caption{ATE RMSE (m) on TUM RGB-D}
\centering
\small
\label{tab:ate_tum}
\setlength{\tabcolsep}{4.5pt}
\providecommand*\rowstyle[1]{\gdef\currentrowstyle{#1}#1\ignorespaces}
\begin{tabular}{>{\gdef\currentrowstyle{}}c!{\hspace{4pt}}>{\currentrowstyle}l!{\hspace{6pt}}>{\currentrowstyle}c>{\currentrowstyle}c>{\currentrowstyle}c>{\currentrowstyle}c>{\currentrowstyle}c>{\currentrowstyle}c>{\currentrowstyle}c>{\currentrowstyle}c>{\currentrowstyle}c!{\hspace{3pt}}>{\currentrowstyle}c>{\currentrowstyle}c>{\currentrowstyle}c}
\toprule
 & & \multicolumn{9}{c}{Sim(3)} & \multicolumn{3}{c}{Average} \\
\cmidrule(lr){3-11} \cmidrule(lr){12-14}
 & & 360 & desk & desk2 & floor & plant & room & rpy & teddy & xyz & Sim(3) & SE(3) & FPS \\
\midrule

\multirow{4}{*}{\rotatebox[origin=c]{90}{\textit{Calibrated}}}
\rowstyle{\color{gray!60}} &
ORB-SLAM3 \cite{orbslam3} & X & 0.016 & X & 0.015 & 0.111 & X & 0.062 & X & 0.009 & X & X & 25.2 \\

\rowstyle{\color{gray!60}}
& DPV-SLAM \cite{dpvslam} & 0.153 & 0.021 & 0.038 & 0.130 & 0.024 & 0.330 & 0.030 & 0.084 & 0.013 & 0.091 & 0.398 & 16.9 \\

\rowstyle{\color{gray!60}}
& DPV-SLAM++ \cite{dpvslam} & 0.175 & 0.021 & 0.037 & 0.137 & 0.020 & 0.236 & 0.026 & 0.090 & 0.012 & 0.084 & 63.971 & 16.8 \\

\rowstyle{\color{gray!60}}
& DROID-SLAM \cite{droidslam} & 0.155 & 0.017 & 0.504 & 0.025 & \textbf{0.017} & 0.071 & 0.057 & 0.045 & 0.040 & 0.103 & 0.258 & 33.6 \\

\rowstyle{\color{gray!60}}
& MASt3R-SLAM \cite{mast3rslam} & \textbf{0.048} & \textbf{0.016} & \textbf{0.024} & \textbf{0.025} & 0.020 & \textbf{0.061} & \textbf{0.023} & \textbf{0.040} & \textbf{0.009} & \textbf{0.030} & \textbf{0.116} & 8.9 \\

\midrule

\multirow{7}{*}{\rotatebox[origin=c]{90}{\textit{Uncalibrated}}}
& DROID-SLAM \cite{droidslam} & 0.205 & 0.029 & 0.091 & 0.045 & 0.057 & 0.599 & 0.059 & 0.071 & 0.013 & 0.130 & 0.241 & 39.9 \\

& MASt3R-SLAM \cite{mast3rslam} & 0.070 & 0.035 & 0.055 & 0.056 & 0.035 & 0.118 & 0.041 & 0.116 & 0.020 & 0.061 & \cellcolor{GreenYellow!50}0.122 & 9.8 \\

& ViSTA-SLAM \cite{vistaslam} & 0.104 & 0.030 & 0.030 & 0.070 & 0.052 & \cellcolor{GreenYellow!50}0.068 & \cellcolor{GreenYellow!50}0.023 & 0.080 & 0.015 & 0.052 & 0.522 & 14.0 \\

& VGGT-SLAM 2.0 \cite{vggtslam2} & 0.060 & 0.026 & \cellcolor{GreenYellow!50}0.027 & 0.043 & 0.033 & 0.070 & 0.029 & \cellcolor{GreenYellow!50}0.038 & 0.020 & 0.038 & 0.249 & 30.5 \\

& AMB3R \cite{amb3r} & \cellcolor{GreenYellow!50}0.052 & \cellcolor{GreenYellow!50}0.021 & 0.030 & \cellcolor{GreenYellow!50}0.034 & \cellcolor{GreenYellow!50}0.032 & 0.072 & 0.025 & 0.047 & \cellcolor{GreenYellow!50}0.012 & \cellcolor{GreenYellow!50}0.036 & 0.231 & 1.04 \\

& DA3-Streaming \cite{da3streaming} & 0.094 & 0.080 & 0.060 & 0.086 & 0.115 & 0.183 & 0.054 & 0.213 & 0.043 & 0.103 & 0.165 & 1.64 \\

& \textbf{DASH-SLAM (Ours)} & \cellcolor{ForestGreen!40}0.046 & \cellcolor{ForestGreen!40}0.017 & \cellcolor{ForestGreen!40}0.020 & \cellcolor{ForestGreen!40}0.025 & \cellcolor{ForestGreen!40}0.028 & \cellcolor{ForestGreen!40}0.063 & \cellcolor{ForestGreen!40}0.019 & \cellcolor{ForestGreen!40}0.027 & \cellcolor{ForestGreen!40}0.009 & \cellcolor{ForestGreen!40}0.028 & \cellcolor{ForestGreen!40}0.078 & 25.3 \\

\bottomrule
\end{tabular}
\end{table*}

\subsection{Implementation Details}
\label{sec:method:impl}

We use the \texttt{DA3NESTED-GIANT-LARGE-1.1} model at a processing resolution of
504~px, with both ViT backbones cast to bfloat16, which reduces peak GPU memory
by approximately 40\% at no measurable accuracy cost. The pose graph is
implemented in GTSAM. All parameters above are shared across experiments.

\section{EXPERIMENTS}

\label{sec:exp}

\subsection{Experimental Setup}
\label{sec:exp:setup}

We evaluate DASH-SLAM against three indoor benchmarks and one challenging field dataset in perceptually-degraded environments. On \textbf{TUM
RGB-D}~\cite{sturm2012tum} we use the nine Freiburg~1 sequences; on
\textbf{Replica}~\cite{straub2019replica} the eight standard scenes; and on
\textbf{7-Scenes}~\cite{7scenes} sequence \texttt{seq-01} of each of the
seven scenes. The field dataset evaluation is reported separately in
Sec.~\ref{sec:exp:uas}.

\textbf{Metrics.} We report the absolute trajectory error (ATE) RMSE under two
alignments. \emph{Sim(3)} is the conventional metric for
monocular SLAM, computed using Umeyama alignment and the ground truth to adjust for a global scale factor and, therefore,
to measure trajectory \emph{shape}. \emph{SE(3)} alignment fits rotation and translation but not
scale, so it additionally penalizes a trajectory that is globally too large or
too small. We also report the average frames-per-second (FPS) rate obtained over each dataset.

\textbf{Baselines.} We compare against 6 uncalibrated systems that, like ours,
receive no intrinsics. In our comparison, we include DA3-Streaming, the streaming reconstruction shipped with
DA3~\cite{da3}, to demonstrate our performance gain over the pure DA3 system. For reference, we also list 5 calibrated systems, which are given the ground-truth intrinsics. Every system is run through the
same harness, with identical dataset loading, association, and scoring, so the
numbers differ only in the tracker. Each system had its metrics computed as its average over 3 runs.

\textbf{Hardware.} All experiments are conducted on a system equipped with an NVIDIA RTX~4000 Ada (20~GB) GPU and an Intel Core Ultra~7 265K CPU. The configuration described in Sec.~\ref{sec:method} is used across all three indoor benchmarks. Additionally, we validate the system on a physical robot equipped with an NVIDIA Jetson AGX Orin (Fig.~\ref{fig:lab_map}).

\textbf{Configurations.} On all indoor datasets, the system was run with $\tau_S = 650~\mbox{px}$, $n_\text{max}=32$, $n_\text{warm}=16$, $n_\text{stride}=4$, $l_\text{diff}=0.8$, $c_{gate}=0.05$, $k_\text{loop}=1.345$. The damping factor was set to $0.0$ for these sequences, corresponding to full scale chaining. For the outdoor aerial sequences, submaps were capped at 16 keyframes, loop closures were made more conservative (with a threshold $l_\text{diff}$ of $0.45$ and a confidence gate $c_{gate}$ of 0.20), and the damping factor was set to $1.0$, relying entirely on DA3's scale, as scale errors tend to compound over very long sequences.

\subsection{TUM RGB-D Dataset}
\label{sec:exp:tum}

Table~\ref{tab:ate_tum} presents per-sequence results. DASH-SLAM attains the lowest
error of all uncalibrated systems on every sequence, with
an average $Sim(3)$ ATE of 0.028~m against 0.036~m for the closest competitor
(AMB3R). The margin widens under $SE(3)$ alignment: 0.078~m against 0.122~m for the
best uncalibrated baseline (MASt3R-SLAM), and 0.231--0.522~m for AMB3R,
VGGT-SLAM and ViSTA-SLAM. Comparison with the calibrated group reveals that DASH-SLAM's $Sim(3)$ average is slightly better than that of the best calibrated system
(MASt3R-SLAM, 0.030~m) and its $SE(3)$ average is also better (0.078~m versus
0.116~m). 

\begin{table}[t]
\centering
\small
\caption{Average ATE RMSE (m) and runtime (FPS) on Replica.}
\label{tab:ate_replica}
\setlength{\tabcolsep}{4.5pt}
\providecommand*\rowstyle[1]{\gdef\currentrowstyle{#1}#1\ignorespaces}
\begin{tabular}{>{\gdef\currentrowstyle{}}c!{\hspace{4pt}}>{\currentrowstyle}l!{\hspace{6pt}}>{\currentrowstyle}c>{\currentrowstyle}c>{\currentrowstyle}c}
\toprule
&  & \textbf{Sim(3)} & \textbf{SE(3)} & \textbf{FPS} \\
\midrule
\multirow{5}{*}{\rotatebox[origin=c]{90}{\textit{Calibrated}}}\rowstyle{\color{gray!60}} & ORB-SLAM3 \cite{orbslam3} & 0.019 & 0.659 & 26.0 \\
\rowstyle{\color{gray!60}} & DPV-SLAM \cite{dpvslam} & 0.899 & 8.449 & 15.0 \\
\rowstyle{\color{gray!60}} & DPV-SLAM++ \cite{dpvslam} & 0.935 & 5.865 & 15.6 \\
\rowstyle{\color{gray!60}} & DROID-SLAM \cite{droidslam} & 0.681 & 0.806 & 74.9 \\
\rowstyle{\color{gray!60}} & MASt3R-SLAM \cite{mast3rslam} & \textbf{0.015} & \textbf{0.093} & 13.6 \\
\midrule
\multirow{7}{*}{\rotatebox[origin=c]{90}{\textit{Uncalibrated}}} & DROID-SLAM \cite{droidslam} & 0.803 & 1.694 & 75.8 \\
& MASt3R-SLAM \cite{mast3rslam} & 0.045 & 0.099 & 17.5 \\
& ViSTA-SLAM \cite{vistaslam} & 0.108 & 0.278 & 21.8 \\
& VGGT-SLAM 2.0 \cite{vggtslam2} & 0.030 & 0.698 & 335.8 \\
& AMB3R \cite{amb3r} & 0.038 & 0.702 & 1.52 \\
& DA3-Streaming \cite{da3streaming} & \cellcolor{GreenYellow!50}0.019 & \cellcolor{GreenYellow!50}0.057 & 2.34 \\
& \textbf{DASH-SLAM (Ours)} & \cellcolor{ForestGreen!40}0.008 & \cellcolor{ForestGreen!40}0.045 & 162.4 \\
\bottomrule
\end{tabular}
\end{table}

\subsection{Replica}
\label{sec:exp:replica}

On Replica (Table~\ref{tab:ate_replica}), DASH-SLAM again achieves the lowest error on every uncalibrated sequence, averaging 0.008~m under $Sim(3)$ alignment and 0.045~m under $SE(3)$. The second-best uncalibrated system, DA3-Streaming, averages 0.019~m and 0.057~m, respectively. Since both systems use the same foundation model, this comparison highlights the contribution of the SLAM layer (i.e. submap batching, anchor-frame scale chaining, pose-graph optimisation, and loop closing) rather than differences in the underlying geometric predictor. These results provide further evidence of DASH-SLAM's generalization across indoor datasets, including synthetic environments.

\begin{table}[t]
\centering
\small
\caption{Average ATE RMSE (m) and runtime (FPS) on 7-Scenes.}
\label{tab:ate_7scenes}
\setlength{\tabcolsep}{4.5pt}
\providecommand*\rowstyle[1]{\gdef\currentrowstyle{#1}#1\ignorespaces}
\begin{tabular}{>{\gdef\currentrowstyle{}}c!{\hspace{4pt}}>{\currentrowstyle}l!{\hspace{6pt}}>{\currentrowstyle}c>{\currentrowstyle}c>{\currentrowstyle}c}
\toprule
&  & \textbf{Sim(3)} & \textbf{SE(3)} & \textbf{FPS} \\
\midrule
\multirow{5}{*}{\rotatebox[origin=c]{90}{\textit{Calibrated}}}\rowstyle{\color{gray!60}} & ORB-SLAM3 \cite{orbslam3} & 0.130 & 0.487 & 24.9 \\
\rowstyle{\color{gray!60}} & DPV-SLAM \cite{dpvslam} & 0.622 & 8.482 & 17.8 \\
\rowstyle{\color{gray!60}} & DPV-SLAM++ \cite{dpvslam} & 0.630 & 4.252 & 17.6 \\
\rowstyle{\color{gray!60}} & DROID-SLAM \cite{droidslam} & 0.341 & 0.433 & 66.6 \\
\rowstyle{\color{gray!60}} & MASt3R-SLAM \cite{mast3rslam}& \textbf{0.049} & \textbf{0.134} & 9.5 \\
\midrule
\multirow{7}{*}{\rotatebox[origin=c]{90}{\textit{Uncalibrated}}} & DROID-SLAM \cite{droidslam} & 0.336 & 0.447 & 67.3 \\
& MASt3R-SLAM \cite{mast3rslam} & 0.066 & 0.150 & 13.2 \\
& ViSTA-SLAM \cite{vistaslam} & \cellcolor{GreenYellow!50}0.055 & 0.429 & 17.0 \\
& VGGT-SLAM 2.0 \cite{vggtslam2} & 0.068 & 0.443 & 87.1 \\
& AMB3R \cite{amb3r} & \cellcolor{ForestGreen!40}0.053 & 0.388 & 0.97 \\
& DA3-Streaming \cite{da3streaming} & 0.082 & \cellcolor{GreenYellow!50}0.093 & 1.74 \\
& \textbf{DASH-SLAM (Ours)} & \cellcolor{ForestGreen!40}0.053 & \cellcolor{ForestGreen!40}0.070 & 82.6 \\
\bottomrule
\end{tabular}
\end{table}

\subsection{7-Scenes}
\label{sec:exp:7scenes}

7-Scenes is the hardest of the three for
foundation-model methods: the sequences are short, hand-held and contain rapid
rotation. DASH-SLAM was best or second-best on five of the seven scenes and tied with
AMB3R for the best uncalibrated $Sim(3)$ average, while being considerably faster (82.6 FPS against 0.97). Under $SE(3)$ the ranking is unambiguous: 0.070~m against 0.093~m for DA3-Streaming and 0.388--0.447~m for AMB3R, VGGT-SLAM and ViSTA-SLAM. 
We report the average ATEs and FPS in Table~\ref{tab:ate_7scenes}.

\begin{table}[t] \centering \small \caption{Estimated trajectory scale and deviation from metric scale.} \label{tab:scale_deviation} \setlength{\tabcolsep}{5pt} \renewcommand{\arraystretch}{1.1} \begin{tabular}{lcc|cc} \toprule & \multicolumn{2}{c|}{\textbf{DASH-SLAM}} & \multicolumn{2}{c}{\textbf{DA3-Streaming}} \\ \textbf{Dataset} & Scale & Deviation & Scale & Deviation \\ \midrule TUM & 1.021 & 2.1\% & 0.871 & 12.9\% \\ Replica & 1.050 & 5.0\% & 1.041 & 4.1\% \\ 7-Scenes & 1.008 & 0.8\% & 0.971 & 2.9\% \\ \bottomrule \end{tabular} 
\end{table}

\subsection{Metric Accuracy}
\label{sec:exp:scale}

Across the three benchmarks, Table~\ref{tab:scale_deviation} shows that the scale factor recovered by DASH-SLAM under $Sim(3)$ alignment remains within 2.1\%, 5.0\%, and 0.8\% of metric scale, respectively, without calibration, an initialisation procedure, or an external reference. This is substantially closer to metric scale than approaches not based on DA3, which do not provide a native mechanism for estimating metric depth. DA3-Streaming, the native implementation of DA3 for long sequences, remains within 3--13\% but does not perform scale chaining across windows. This comparison suggests that metric depth alone is not sufficient: the per-batch scales must also be reconciled, at least for smaller sequences, as provided by the anchor-frame chaining described in Sec.~\ref{sec:method}.

\begin{table}[t]
\centering
\small
\caption{UAS ATE RMSE (m) with Sim(3) alignment}
\label{tab:ate_uas}
\setlength{\tabcolsep}{5pt}
\renewcommand{\arraystretch}{1.1}
\begin{tabular}{lcccc}
\toprule
& Fyllin. & Hornb. & Frozen& Campus \\
& & & lake & fog \\
\midrule
{\footnotesize ORB-SLAM3}\cite{orbslam3}
    & $\times$
    & $\times$
    & \cellcolor{ForestGreen!40}6.84
    & \cellcolor{ForestGreen!40}1.68 \\

{\footnotesize VGGT-SLAM2.0}\cite{vggtslam2}
    & \cellcolor{GreenYellow!50}196.93
    & \cellcolor{GreenYellow!50}148.90
    & 76.06
    & 46.08 \\

{\footnotesize \textbf{DASH-SLAM}}
    & \cellcolor{ForestGreen!40}64.19
    & \cellcolor{ForestGreen!40}75.84
    & \cellcolor{GreenYellow!50}74.35
    & \cellcolor{GreenYellow!50}2.66 \\

\bottomrule
\end{tabular}
\end{table}

\begin{figure*}[!t]
\centering
\includegraphics[width=\textwidth]{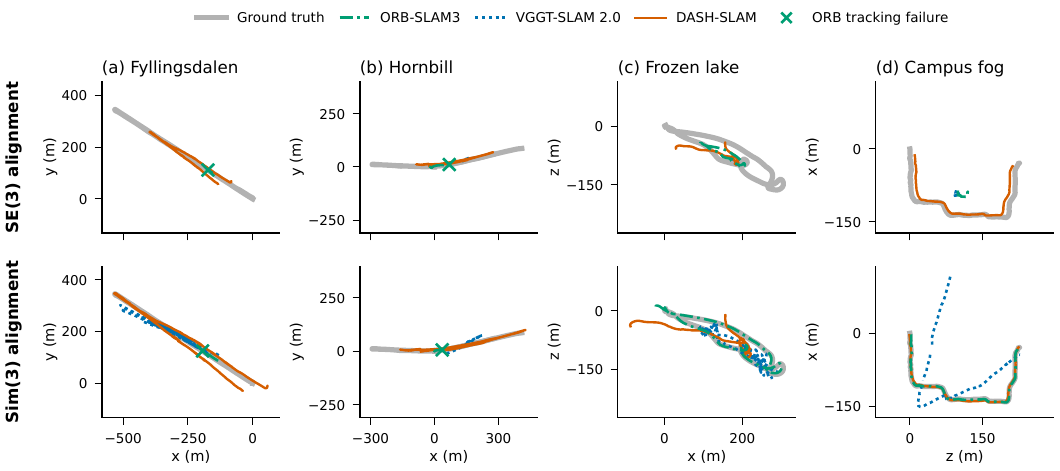}
\caption{Ground truth and estimated UAS trajectories after Sim(3) and SE(3) alignment.
Crosses mark the last ORB-SLAM3 poses before tracking failures. }
\label{fig:uas_all_trajectories}
\end{figure*}

\subsection{Unified Autonomy Stack (UAS) Dataset}
\label{sec:exp:uas}

As the three benchmarks above cover indoor environments and traverse at most a few
tens of meters, we use four sequences of the NTNU-ARL
UAS dataset to test our approach in much more difficult regimes: two tunnel inspections (\texttt{fyllingsdalen}, \texttt{hornbill}), a frozen lake and a foggy campus flight, whose ground-truth paths are 659--1427~m long. The camera is fisheye, the scenes are textureless or fog-obscured over long stretches, and the flights are an order of magnitude longer. 

Table~\ref{tab:ate_uas} compares DASH-SLAM with VGGT-SLAM and ORB-SLAM3, the latter provided with the fisheye camera calibration. DASH-SLAM achieves ATEs of 64.19~m and 75.84~m under $Sim(3)$ alignment on the \texttt{fyllin} and \texttt{hornbill} sequences, respectively, substantially outperforming VGGT-SLAM, which obtains 196.93~m and 148.90~m. ORB-SLAM3 loses tracking on both sequences; it re-initialises a new map and retains only the largest one, which on \texttt{hornbill} covers only 5\% of the route. On \texttt{frozen lake}, however, ORB-SLAM3 tracks the sequence to its end and achieves a substantially lower ATE, probably due to the abundance of visual texture, although highly self-similar, on the lake's icy surface. DASH-SLAM and VGGT-SLAM obtain comparable errors on this sequence, at 74.35~m and 76.06~m, respectively.
Finally, \texttt{campus\_fog} presents a particularly challenging trajectory. None of the systems successfully reconstructs the complete sequence. Nevertheless, ORB-SLAM3 and DASH-SLAM achieve substantially lower ATEs when evaluation is restricted to the first half of the trajectory, at 1.68~m and 2.66~m, respectively, whereas VGGT-SLAM still exhibits a high ATE of 46.08~m.
Since comparing ATE values can be misleading, we provide visualizations of the estimated trajectories versus the ground truth with both Sim(3) and SE(3) alignment in Figure~\ref{fig:uas_all_trajectories}.

\textbf{Scale chaining does not transfer unchanged to kilometre-scale trajectories.}
The indoor configuration chains every boundary ratio ($g = 0$). On the UAS sequences, however, this setting degrades the trajectory by 29--58\% and drives the recovered scale to 0.10--0.38. This degradation results from the length of the trajectories: although the scale estimates at individual aerial boundaries have accuracy comparable to those observed indoors, their errors compound over hundreds of boundaries. Moreover, under low or ambiguous texture conditions with many self-similar views, such as in tunnels, loop closure retrieval must be more conservative, as a permissive threshold can lead to false-positive loop closures. The aerial results therefore use $g = 1$, which disables scale chaining and retains each submap's independently estimated metric scale, together with shorter submaps (16 keyframes). A more sophisticated scale model that accounts for these different operating conditions could potentially address this limitation.

\section{CONCLUSION}

We presented DASH-SLAM, a monocular SLAM system that integrates metric-depth predictions through anchor-frame scale chaining and a shared-keyframe graph. Depth-derived corrections enable scale accuracy and consistency across odometry, loop closure measurements, and map geometry. Across three indoor benchmarks, DASH-SLAM achieves the lowest reported metric trajectory error among the evaluated methods, while reaching real-time speeds (25.3 to 162.4 FPS). Our technique also obtains lower error than a comparable GFM-based method across four challenging sequences in perceptually-degraded environments, although substantial residual errors remain. These results point to a promising future for metric foundation models in SLAM and motivate further research toward robust operation in challenging real-world environments beyond indoor benchmarks.

\bibliographystyle{IEEEtranN}
\bibliography{bib/references}

\end{document}